\documentclass[lettersize,journal]{IEEEtran}
\usepackage[inkscapelatex=false]{svg}
\usepackage{amsmath,amsfonts}
\usepackage{algorithmic}
\usepackage{algorithm}
\usepackage{array}
\usepackage{textcomp}
\usepackage{stfloats}
\usepackage{url}
\usepackage{verbatim}
\usepackage{graphicx}
\usepackage{epstopdf}
\usepackage{cite}
\usepackage{amsmath,amsfonts}
\usepackage{multirow}
\usepackage{makecell}
\usepackage{subcaption}
\usepackage{booktabs}
\usepackage{tcolorbox}
\usepackage[dvipsnames]{xcolor}
\usepackage{hyperref}[colorlinks,linkcolor=blue]
\usepackage{cleveref}

\definecolor{kitti_car}{RGB}{100, 150, 245}
\definecolor{kitti_bicycle}{RGB}{100, 230 ,245}
\definecolor{kitti_motorcycle}{RGB}{30, 60, 150}
\definecolor{kitti_truck}{RGB}{80, 30, 180}
\definecolor{kitti_other-vehicle}{RGB}{0, 0, 255}
\definecolor{kitti_person}{RGB}{255, 30, 30}
\definecolor{kitti_bicyclist}{RGB}{255, 40, 200}
\definecolor{kitti_motorcyclist}{RGB}{150, 30, 90}
\definecolor{kitti_road}{RGB}{255, 0, 255}
\definecolor{kitti_parking}{RGB}{255, 150, 255}
\definecolor{kitti_sidewalk}{RGB}{75, 0, 75}
\definecolor{kitti_other-ground}{RGB}{175, 0, 75}
\definecolor{kitti_building}{RGB}{255, 200, 0}
\definecolor{kitti_fence}{RGB}{255, 120, 50}
\definecolor{kitti_vegetation}{RGB}{0, 175, 0}
\definecolor{kitti_trunk}{RGB}{135, 60, 0}
\definecolor{kitti_terrain}{RGB}{150, 240, 80}
\definecolor{kitti_pole}{RGB}{255, 240, 150}
\definecolor{kitti_trafficsign}{RGB}{255, 0, 0}

\definecolor{nus_others}{RGB}{128, 128, 128}       
\definecolor{nus_barrier}{RGB}{255, 120, 50}
\definecolor{nus_bicycle}{RGB}{255, 192, 203}
\definecolor{nus_bus}{RGB}{255, 255, 0}
\definecolor{nus_car}{RGB}{0, 255, 127}
\definecolor{nus_construction_vehicle}{RGB}{0, 255, 255}
\definecolor{nus_motorcycle}{RGB}{200, 180, 0}
\definecolor{nus_pedestrian}{RGB}{255, 0, 0}
\definecolor{nus_traffic_cone}{RGB}{255, 240, 150}
\definecolor{nus_trailer}{RGB}{135, 60, 0}
\definecolor{nus_truck}{RGB}{160, 32, 240}
\definecolor{nus_driveable_surface}{RGB}{255, 0, 255}
\definecolor{nus_other_flat}{RGB}{139, 137, 137}
\definecolor{nus_sidewalk}{RGB}{75, 0, 75}
\definecolor{nus_terrain}{RGB}{150, 240, 80}
\definecolor{nus_manmade}{RGB}{230, 230, 250}
\definecolor{nus_vegetation}{RGB}{0, 175, 0}

\begin{document}

\title{Explicit Semantics and Uncertainty Guided Sparse Learning for Efficient 3D Occupancy Prediction}

\author{Hanlin Wu\textsuperscript{\href{https://orcid.org/0000-0002-9640-9315}{\includegraphics[scale=0.06]{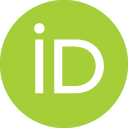}}},~\IEEEmembership{Student Member,~IEEE}, Pengfei Lin\textsuperscript{\href{https://orcid.org/0000-0002-8609-9341}{\includegraphics[scale=0.06]{figures/ORCIDiD.png}}},~\IEEEmembership{Member,~IEEE}, Ehsan Javanmardi{\href{https://orcid.org/0000-0003-0337-115X}{\includegraphics[scale=0.06]{figures/ORCIDiD.png}}},~\IEEEmembership{Member,~IEEE}, Naren Bao{\href{https://orcid.org/0000-0001-9602-8830}{\includegraphics[scale=0.06]{figures/ORCIDiD.png}}},~\IEEEmembership{Member,~IEEE}, Bo Qian\textsuperscript{\href{https://orcid.org/0000-0002-6964-220X}{\includegraphics[scale=0.06]{figures/ORCIDiD.png}}},~\IEEEmembership{Member,~IEEE}, Hao Si\textsuperscript{\href{https://orcid.org/0009-0000-4348-8702}{\includegraphics[scale=0.06]{figures/ORCIDiD.png}}}, Manabu Tsukada\textsuperscript{\href{https://orcid.org/0000-0001-8045-3939}{\includegraphics[scale=0.06]{figures/ORCIDiD.png}}},~\IEEEmembership{Member,~IEEE}
\thanks{Hanlin Wu,  Pengfei Lin, Ehsan Javanmardi, Naren Bao, Bo Qian, Hao Si and Manabu Tsukada are with the Graduate School of Information Science and Technology, The University of Tokyo, Tokyo, 113-8657, Japan. (e-mail: $\{$hanlinwu, linpengfei0609, ejavanmardi, naren$\}$@g.ecc.u-tokyo.ac.jp, boqian@ieee.org, $\{$naren, si-hao, mtsukada$\}$@g.ecc.u-tokyo.ac.jp)}%

}

\markboth{Journal of \LaTeX\ Class Files,~Vol.~18, No.~9, September~2020}%
{Shell \MakeLowercase{\textit{et al.}}: A Sample Article Using IEEEtran.cls for IEEE Journals}



\maketitle


\begin{abstract}
    3D semantic occupancy prediction has emerged as a critical perception task for autonomous driving due to its ability to offer voxel-level semantic and geometric understanding of the environment. However, such a refined representation for large-scale scenes incurs prohibitive computation, posing a significant challenge to practical real-time deployment. To address this, we propose SUGOcc, an explicit semantics and uncertainty guided sparse learning framework for efficient occupancy prediction, which exploits the inherent sparsity of 3D scenes to reduce redundant computation while maintaining geometric and semantic integrity. Specifically, we first utilize semantic and uncertainty priors to suppress image projections from free space while employing explicit unsigned distance encoding to enhance geometric consistency, thereby producing a structurally sparse representation. Secondly, we introduce a cascade sparse completion module to enable efficient coarse-to-fine reasoning over the sparse representation via hyper cross sparse convolution, generative upsampling and adaptive pruning. Finally, we propose an object contextual representation (OCR) based mask decoder that refines the voxel-wise predictions through lightweight query-context interactions, thereby avoiding expensive attention operations over volumetric features. Extensive experiments on SemanticKITTI and Occ3D-Nuscenes benchmark demonstrate that the proposed approach outperforms the baselines, achieving notable improvements in both accuracy and efficiency across datasets. Code and models are available at \hyperlink{https://github.com/tlab-wide/SUGOcc}{https://github.com/tlab-wide/SUGOcc}.
\end{abstract}


\section{Introduction}
\IEEEPARstart{T}{he} 3D semantic occupancy prediction has emerged as a fundamental perception task for scene understanding in autonomous driving, thanks to its ability to jointly estimate voxel-level geometric and semantic states in 3D space \cite{xu2024survey}, \cite{11386780}, \cite{10694710}. Differing from conventional perception paradigms like image segmentation \cite{guo2022segnext, chen2017deeplab}, bounding-box detection \cite{yan2018second, lang2019pointpillars}, and Bird-Eye-View (BEV) map \cite{li2024bevformer, liu2022bevfusion}, 3D semantic occupancy prediction produce dense and spatially consistent 3D scene representations which are beneficial for downstream tasks such as motion planning \cite{wu2022deep} and other safety-critical decision-making \cite{8848796} in real-time autonomous systems.

Despite these advantages, 3D semantic occupancy prediction poses a critical computational bottleneck for real-time deployment. Most existing approaches follow a two-stage pipeline, where multi-view image features are first lifted into a 3D volume via view transformation, typically based on Lift-Splat-Shoot (LSS) \cite{philion2020lift} or transformer-driven BEV mapping \cite{li2024bevformer}, and then processed by dense 3D Convolutional Neural Networks (CNNs) or transformer decoders. While effective, this design leads to substantial memory overhead and inference latency due to the massive intermediate feature space. Even a 3D volume with moderate spatial resolution contains hundreds of thousands of locations.

\begin{figure}
    \centering
    \includegraphics[width=0.99\linewidth]{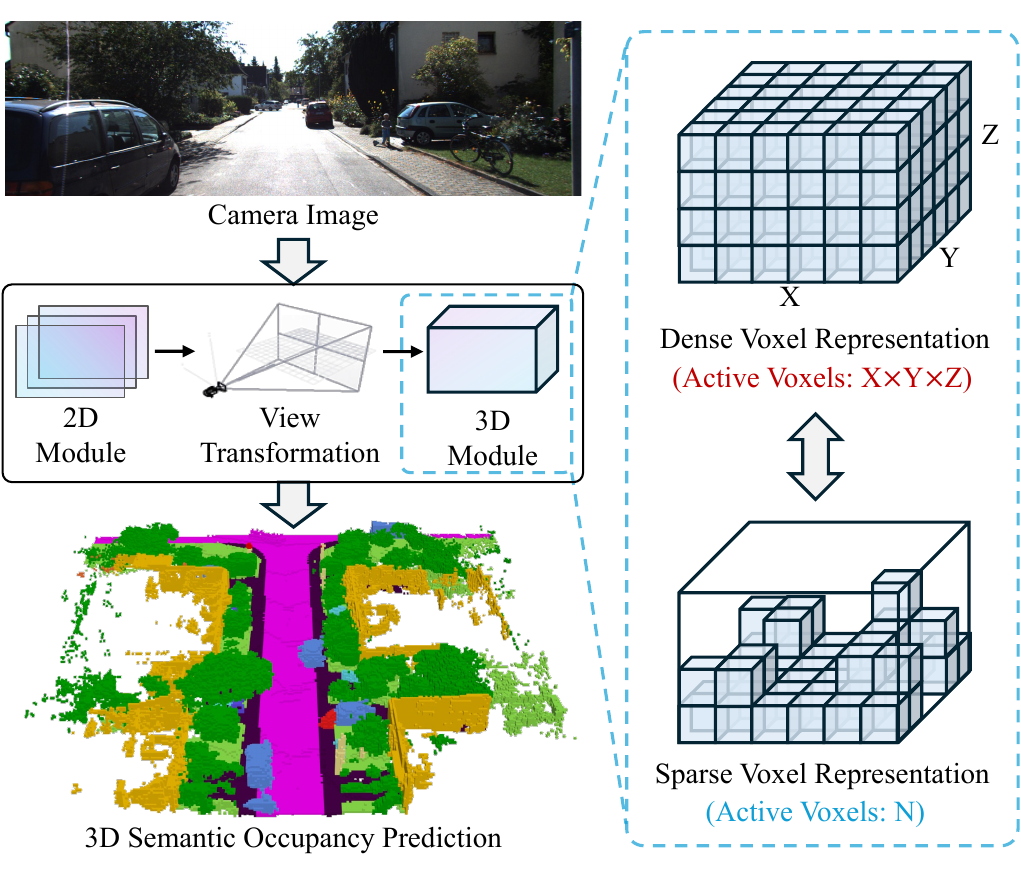}
    \captionof{figure}{Illustration of the inherent sparsity of 3D semantic occupancy prediction.}
    \label{fig:sparse}
\end{figure}

At the core of the inefficiency lies a fundamental mismatch between dense computation and the inherently sparse nature of real-world 3D environments. In typical autonomous driving scenes, only a small fraction of voxels correspond to physical objects, while the overwhelming majority represent free space. Nevertheless, existing occupancy prediction methods allocate equal computational resources to all voxels, i.e., occupied or empty, resulting in substantial redundancy. Consequently, a significant portion of the computational budget is wasted on voxels that contribute little to perception quality. This structural inefficiency presents a major roadblock to deploying semantic occupancy prediction in real-time autonomous driving systems.

Driven by this sparsity imbalance, recent works like SparseOcc \cite{tang2024sparseocc} and Pasco \cite{cao2024pasco} have explored the promise of introducing sparse representations into the occupancy prediction pipeline by exploiting the natural sparsity that emerges from LiDAR scans or from the geometric misalignment between camera rays and voxel centers during the Lift-Splat-Shoot (LSS) projection. While this reduces the number of active voxels, the resulting sparsity is incidental rather than principled, which stems from projection geometry rather than the semantic structure of the scene. As a result, numerous voxels corresponding to free space or uncertain projections remain within the sparse volume, continuing to consume computational resources with limited perceptual value.

More critically, existing sparse methods \cite{liu2024fully} \cite{wang2024opus} often lack high-level image priors when deciding which voxels to retain or discard. Depth distributions are typically predicted, but only used as projection weights during view transformation, which are often sharply peaked around visible surfaces, providing limited expressiveness along the camera ray. Furthermore, high-level semantic priors are entirely decoupled from the lifting process and are not exploited to suppress semantically implausible or free-space projections \cite{tang2024sparseocc}. As a result, unreliable or semantically implausible projections contaminate the 3D feature space, undermining the intended benefits of sparsity. The pipeline still ends up processing many voxels that offer little to no geometric insight. This not only degrades the representational quality of sparse volumes, but also introduces computational waste, posing a major barrier to real-world deployment of sparse occupancy prediction systems.

To address the limitations of existing 3D semantic occupancy prediction methods, we propose SUGOcc, an efficient semantic occupancy prediction framework that leverages explicit semantics and uncertainty guided sparse learning to reduce computational cost while enhancing geometric and semantic consistency. To be specific, in order to obtain a cleaner and more informative sparse 3D representation, we introduce an explicit semantics and uncertainty guided LSS module to selectively project image features into 3D space based on the semantic and depth uncertainty priors rather than indiscriminately lifting all image features. In addition, to efficiently perform the coarse-to-fine geometric and semantic reconstruction over the sparse 3D representations, we design a cascade sparse completion network, which progressively expands active voxels via generative upsampling, efficiently refines structural voxels using hyper-cross sparse convolutions, and adaptively prunes low-confidence regions to tightly control computational overhead. Furthermore, for final semantic occupancy predictions, we propose a lightweight OCR-based mask decoder that derives object contextual representations from sparse voxel features and restricts query interactions to this compact context, avoiding dense voxel-wise attention operations. A global OCR embedding is also maintained and updated via exponential moving average accumulation, providing a global semantic prior with negligible overhead. 


In summary, the contributions of this work are threefold:
\begin{itemize}
    \item We introduce an explicit semantics and uncertainty guided LSS module to select informative image projections and incorporate an explicit distance encoding, producing a geometrically and semantically coherent sparse 3D representation with controllable active voxels.
    \item We design a cascade sparse completion network that enables efficient coarse-to-fine occupancy completion over sparse voxels via hyper-cross sparse convolution, generative upsampling and adaptive pruning.
    \item We propose an OCR-based mask decoder to efficiently refine semantic occupancy prediction, which aggregate object context from sparse volume features, maintain global context and perform only query-context interactions, avoiding multi-scale and dense voxel-wise attention.
\end{itemize}

Extensive experiments on SemanticKITTI and Occ3D-Nuscenes dataset demonstrate that the proposed approach achieves superior performance over baselines, yielding a $7.34\%$ accuracy improvement and a $57.8\%$ efficiency enhancement on SemanticKITTI, with notable performance gains observed on Occ3D-Nuscenes.

\section{Related Work}

\subsection{2D-to-3D View Transformation}
Transforming image features into structured 3D representations is a fundamental component of vision-based 3D perception in autonomous driving. Existing methods can be broadly divided into explicit and implicit view transformation paradigms. In explicit view transformation, the pioneering LSS \cite{philion2020lift} introduced a dominant paradigm: it predicts a discrete depth distribution for each pixel and explicitly lifts image features into a 3D frustum. Due to its geometric interpretability and efficiency, LSS has served as the foundation for a wide range of BEV-based detection and perception systems \cite{reading2021categorical}, \cite{huang2021bevdet}, \cite{huang2022bevdet4d}. Subsequent extensions such as BEVDepth \cite{li2023bevdepth} further enhanced this paradigm by incorporating camera-aware modeling and explicit depth supervision.

In contrast, implicit view transformation learns pixel-to-3D correspondences directly through attention-based mechanisms without relying on explicit depth estimations. Representative methods include BEVFormer \cite{li2024bevformer}, which performs deformable cross-attention between BEV queries and image features. Building on this, SurroundOcc \cite{Wei_2023_ICCV} extends this formulation from BEV representations to semantic occupancy predictions. Together, explicit and implicit view transformation paradigms provide two dominant approaches for constructing 3D representations from 2D images and form the foundation of modern BEV and voxel-based perception systems.

\subsection{3D Semantic Occupancy Prediction}
3D semantic occupancy prediction aims to jointly infer the occupancy status and semantic category of each voxel in a 3D scene, offering a unified representation of geometry and semantics. This task was first introduced by SSCNet \cite{song2017semantic}, which predicts complete 3D semantic scenes from a single-view depth image in indoor environments. With the advent of large-scale datasets such as SemanticKITTI \cite{behley2019iccv}, \cite{behley2021ijrr}, SSCBench \cite{li2023sscbench}, and Occ3D \cite{tian2024occ3d}, semantic occupancy prediction has been extended to complex outdoor scenarios in autonomous driving.

Early works on semantic occupancy prediction predominantly relied on LiDAR-based inputs due to their accurate geometric measurements. Representative approaches such as LMSCNet \cite{9320442}, S3CNet \cite{pmlr-v155-cheng21a} and JS3CNet \cite{yan2021sparse} demonstrated the effectiveness of volumetric reasoning for reconstructing dense 3D semantics from sparse point clouds.

More recently, vision-centric semantic occupancy prediction has gained increasing attention due to its strong potential and lower deployment cost in autonomous driving. Methods like MonoScene \cite{Cao_2022_CVPR}, SurroundOcc \cite{Wei_2023_ICCV}, OccFormer \cite{Zhang_2023_ICCV}, VoxFormer \cite{10203337}, OccNeRF \cite{11003427}, SparseOcc \cite{tang2024sparseocc}, ALOcc \cite{chen2025alocc} and ProtoOcc \cite{kim2025protoocc}, have progressively advanced the capability of vision-based models to infer high-quality 3D semantic occupancy predictions from multi-view images. Despite these advances, vision-based semantic occupancy prediction remains fundamentally limited by depth ambiguity, noisy or inconsistent view transformation, and the high computational overhead required for real-time deployment.

\begin{figure*}[htbp]
    \centering
    \includegraphics[width=0.99\linewidth]{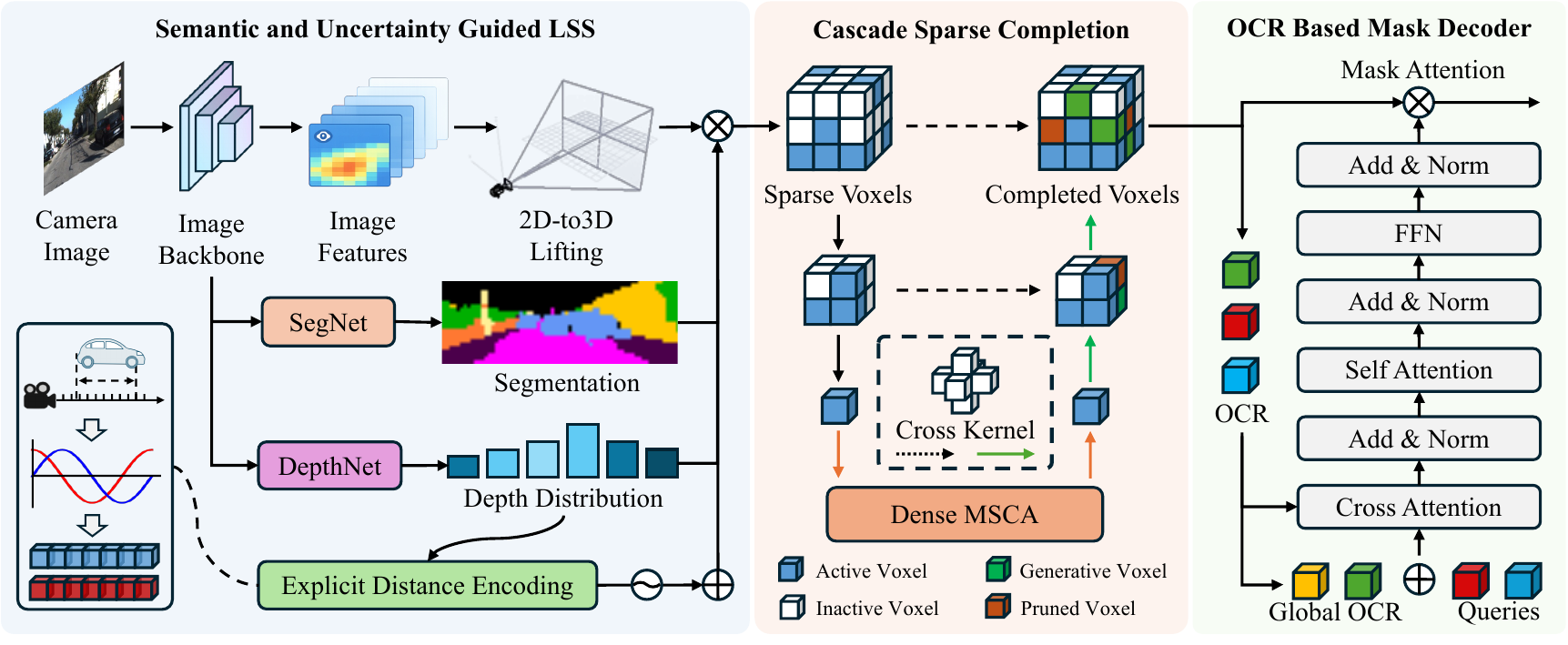}
    \captionof{figure}{Overview of the proposed SUGOcc. At first, image features extracted from a single-frame camera input are selectively lifted into 3D space using semantic and depth uncertainty priors, yielding a sparse and structurally coherent initialization with explicit distance encoding. Then, an efficient cascade sparse completion network progressively reconstructs geometry and semantics while tightly controlling computational cost through generative upsampling and soft pruning. Finally, an OCR-based mask decoder further refines predictions by restricting attention to compact object contextual representations.}
    \label{fig:pipeline}
\end{figure*}

\subsection{Sparse Learning for 3D Perception}
Offering an efficient paradigm by operating only on active positions and avoiding redundant computation inherent in dense representations, sparse learning \cite{liu2015sparse}, \cite{graham2017submanifold}, \cite{choy20194d} is widely used by LiDAR-based perception systems, where the natural sparsity of point clouds aligns well with the sparse computational model. Representative approaches \cite{yan2018second}, \cite{gwak2020generative} and subsequent variants \cite{liu2022spatial}, \cite{chen2022focal} demonstrated that sparse 3D architectures can effectively support large-scale real-time perception. 

However, directly transferring sparse learning to camera-based perception is non-trivial due to the inherently dense nature of multi-view image inputs. To bridge this gap, recent works have explored introducing sparsity into the vision-based 3D perception. For instance, SparseBEV \cite{liu2023sparsebev} proposes sparse BEV representations with token selection strategy for 3D object detection, while PointBEV \cite{chambon2024pointbev} lifts multi-view image features into sparse point-based representations and performs 3D reasoning on sampled keypoints. Nevertheless, the sparsity patterns in these approaches are largely driven by heuristic sampling or attention-based relevance estimation, and lack explicit geometric and semantic grounding, which is crucial for semantic occupancy prediction.

\section{Methodology}
In this section, we introduce the proposed framework for real-time 3D semantic occupancy prediction that leverages explicit semantics and uncertainty guided sparse learning.

\subsection{Problem Formulation}
We formulate the task as predicting a voxel-wise semantic occupancy field of a 3D driving scene from a set of calibrated camera images. Given input images $I\in\mathbb{R}^{N\times H \times W \times 3}$ captured by $N$ cameras where $H$ and $W$ denote the image height and width, respectively. Each camera is associated with an intrinsic matrix $K\in R^{3\times3}$ and an extrinsic pose $T\in SE(3)$. The objective is to infer a 3D semantic occupancy grid $Occ\in\mathbb{R}^{X\times Y \times Z}$, where $X$, $Y$, and $Z$ denote the spatial dimensions of the voxel grid along the longitudinal, lateral, and vertical axes. Each voxel in $Occ$ represents the semantic category of its corresponding spatial location, including free space as well as object categories like car, pedestrian, and vegetation.

\subsection{Overview}
The proposed framework follows a three-stage pipeline for 3D semantic occupancy prediction. Given the input images, we first apply an \textbf{Explicit Semantics and Uncertainty Guided LSS} to obtain a clean and structurally plausible initial 3D sparse volume features. Instead of indiscriminately projecting all image features into the voxel grid, this module leverages semantic priors and depth-based uncertainty estimation to filter out unreliable projections and suppress high-confidence free-space regions before lifting into 3D domain. Then, an efficient \textbf{Cascade Sparse Completion Network} processes the filtered 3D sparse volume features using Minkowski sparse convolutions. This network performs 3D reasoning only on active voxels and progressively reconstructs geometric structures through generative upsampling, while adaptively pruning semantically implausible locations to tightly control computational overhead. Finally, an \textbf{OCR-based Mask Decoder} aggregates global semantic context from the sparse voxel features and refines voxel-wise predictions via a query–context interaction mechanism. By restricting attention operations to compact contextual representations, the proposed framework produces the final dense semantic occupancy field with significantly reduced computational overhead.

\begin{figure}
    \centering
    \includegraphics[width=0.99\linewidth]{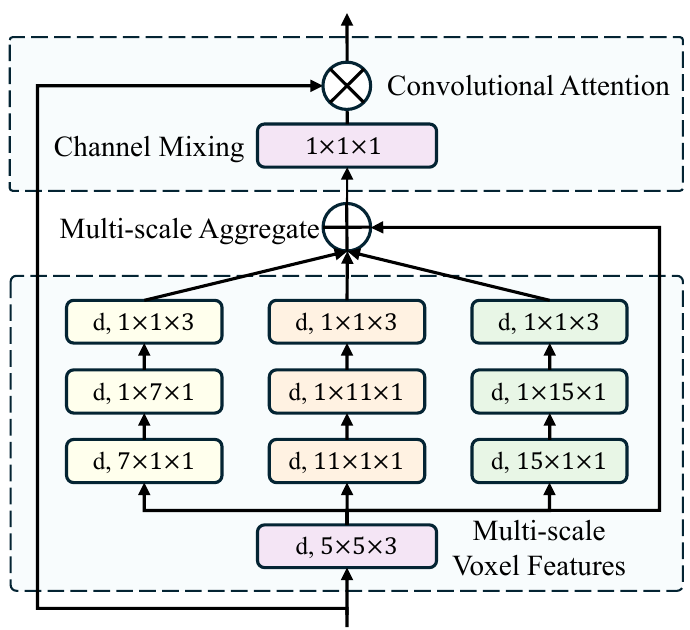}
    \captionof{figure}{Illustration of the multi-scale convolutional attention, which is consist of multi-scale decomposed depth-wise convolution for context aggregation, a point-wise convolution for channel mixing and a convolutional attention for residual operation.}
    \label{fig:msca}
\end{figure}

\subsection{Explicit Semantics and Uncertainty Guided LSS}
The view transformation step plays a crucial role in determining the quality of the initial 3D feature volume. However, standard LSS pipelines indiscriminately lift all image features into the voxel space, regardless of their semantic relevance or geometric reliability. As a result, regions corresponding to free space are still projected, introducing substantial noise and redundancy in the 3D domain. To address this issue, we propose an explicit semantics and uncertainty guided LSS mechanism that selectively removes redundant projections in view transformation and enhances geometric consistency via explicit distance encoding.

Given the input images, we extract the multi-scale image features using a shared image backbone (e.g., ResNet-50), followed by feature pyramid fusion to obtain a downsampled image feature map $F\in \mathbb{R}^{H_d\times W_d \times C}$, which serves as the input to the LSS module. Based on this image feature map, we perform two auxiliary predictions, a semantic segmentation $F_{sem}\in \mathbb{R}^{H_d\times W_d \times S}$ and a depth distribution $F_{depth}\in \mathbb{R}^{H_d\times W_d \times D}$, where $S$ and $D$ denote the number of semantic classes and discretized depth bins, respectively.

For each pixel $(h, w)$, we obtain the semantic probability $P_{sem}(h, w) \in \mathbb{R}^{S}$ and the depth probability distribution $P_{depth}(h, w) \in \mathbb{R}^{D}$ via a softmax operation. The semantic segmentation provides a measure of whether a pixel corresponds to a physically meaningful structure in the 3D scene.
Assuming that class $0$ represents free space, we define the non-empty probability of a pixel as
\begin{equation}
    P_{nonempty}(h,w)=1-P_{sem}(h,w,0)
\end{equation}

Furthermore,  we quantify the geometric uncertainty of projected depths along each camera ray using the cumulative distribution of the predicted depth distribution. To be specific, for each pixel $(h, w)$ and depth bin $d$, we define a cumulative depth confidence as
\begin{equation}
    P_{uncertain}(h,w,d)=\sum_{i=0}^{d} P_{depth}(h,w,i)
\end{equation}
which serves as a proxy for depth uncertainty. A low cumulative probability indicates a foreground-dominated depth bin, whereas a high value suggests a more ambiguous geometric estimate.

Together with the semantic prior, these two probabilities enable us to identify and discard ambiguous or low-quality image positions before lifting the image features into the 3D voxel space. We construct a binary mask that suppresses pixels with a high likelihood of belonging to free space or exhibiting unreliable geometric estimates.
\begin{equation}
Mask(h,w,d) =
\begin{cases}
1, & \text{if } P_{\text{nonempty}}(h,w) > \tau_s \text{ and } \\
   & \quad  P_{\text{uncertain}}(h,w,d) > \tau_d, \\
0, & \text{else}.
\end{cases}
\end{equation}
where $\tau_s$ and $\tau_d$ are predefined thresholds controlling semantic relevance and geometric reliability, respectively.

During view transformation, voxel features are computed only from masked-valid pixels, resulting in a significantly cleaner and more structurally coherent 3D initialization. By suppressing free-space and ambiguous projections at the view transformation stage, the resulting 3D sparse volume features provide a more reliable foundation for subsequent sparse 3D reasoning.

Moreover, unlike the original LSS formulation that aggregates features using the depth distribution probabilities as interpolation weights which often leads to geometric blurring under peak depth distribution, we introduce an explicit unsigned distance encoding to preserve geometric continuity along each camera ray. Instead of probabilistically spreading features across depth bins, we anchor features at the expected depth and encode their relative distance to each depth hypothesis. Specifically, given the depth probability distribution $P_{depth}(h,w,d)$, we estimate the expected depth as
\begin{equation}
    E_{depth}(h,w) = \sum_{i=0}^Di\cdot{P_{depth}(h,w,i)}
\end{equation}
where $d$ and $i$ denote discrete depth bin indices.

We then compute the unsigned distance between each depth bin and the expected depth as
\begin{equation}
    \Delta_{dist}(h,w,d) = \lvert d-E_{depth}(h,w) \rvert
\end{equation}
This relative distance is embedded using a sinusoidal positional encoding, which is defined as
\begin{equation}
    \begin{split}
        PE_{(h,w,d,2i)} &= \mathrm{sin}(\Delta_{dist}(h,w,d)/T^{2i/C}) \\
        PE_{(h,w,d,2i+1)} &= \mathrm{cos}(\Delta_{dist}(h,w,d)/T^{2i/C})
    \end{split}
\end{equation}
where $T$ is a temperature hyperparameter controlling the frequency scale.

Finally, the explicit distance encoding is added to the original image feature to form the lifting features, which is defined as
\begin{equation}
    F^{\prime}(h,w,d) = F(h,w) + PE_{(h,w,d)}
\end{equation}

Based on the explicit distance embedded image features and the explicit semantics and uncertainty guided mask, we obtain an initial sparse volume feature $V_0\in\mathbb{R}^{X_{r}\times Y_{r}\times Z_{r}\times C}$, where $X_{r}, Y_{r}, Z_{r}$ denote the spatial dimensions at resolution $r$ with respect to the full voxel grid.

\subsection{Cascade Sparse Completion}
To efficiently perform 3D occupancy completion on the sparse volume features produced by the explicit semantics and uncertainty guided LSS module, we introduce a cascade sparse completion module, built upon a sparse generative 3D U-Net \cite{gwak2020generative}.

To be specific, the input sparse volume feature $V_0$ is processed by a sparse encoder consisting of submanifold sparse convolutions and sparse downsampling layers, producing multi-scale sparse volume features $V_i$, $i\in\{0,1,2,3\}$ at resolution $s\in \{r, r/2, r/4, r/8\}$. The coarsest feature volume $V_3$ at resolution $r/8$ serves as the bottleneck representation. To enhance multi-scale geometric context and enable geometry generation beyond the input sparse manifold, we employ a lightweight dense multi-scale convolutional attention (MSCA) module \cite{guo2022segnext} on this bottleneck representation, as illustrated in Fig. \ref{fig:msca}. The MSCA-enhanced bottleneck feature is computed as
\begin{equation}
    V_3^{\prime} = V_3\otimes Conv_{1\times1\times1}(\sum_{i=0}^3Scale_i(DW\text{-}Conv(V_3)))
\end{equation}
where $DW\text{-}Conv$ denotes depth-wise convolution \cite{chollet2017xception}, $Scale_i$ denotes the $i$-th multi-scale branches, $Conv_{1\times1\times1}$ denotes point-wise convolution, and $\otimes$ indicates element-wise multiplication.

On top of the bottleneck feature, we employ a cascade sparse generative decoder. At each decoding stage, features from the coarser scale are progressively upsampled via generative transpose convolution to activate new voxels. Following the U-Net paradigm, skip connections from the corresponding encoder stages are fused to preserve fine-grained geometric structures. In addition, we attach a lightweight semantic occupancy predictor at each stage to estimate a proxy semantic occupancy map $P_{occ,i}$, which provides informative supervision across multiple scales.

Guided by the predicted proxy semantic occupancy map, we perform a soft sparse pruning after each decoding stage to maintain sparsity for subsequent stages while preserving meaningful structures. Instead of hard pruning based on binary occupancy or semantic masks as commonly adopted in prior work, we only discard voxels with high confidence of being empty. Formally, we retain voxels whose predicted non-empty probability exceeds a threshold, which is defined as
\begin{equation}
    V_i^{\prime} = \{x\in V_i|P_{occ,i}(x)>\tau_p\}
\end{equation}
where $\tau_p$ denotes the sparse pruning threshold.

\begin{figure}
    \centering
    \includegraphics[width=0.9\linewidth]{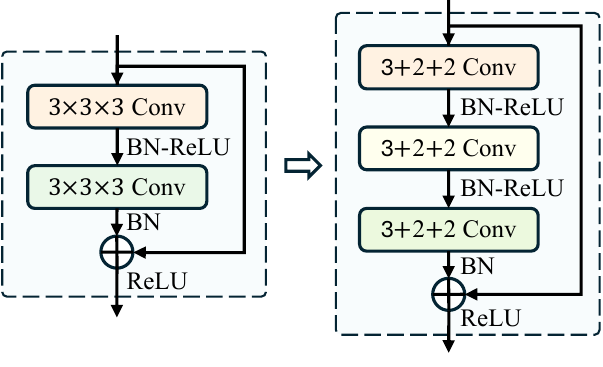}
    \captionof{figure}{Illustration of the proposed hyper cross residual block. Compared with the conventional residual block using two stacked $3\times3\times3$ sparse convolutions, hyper cross residual block apply a three-layer $3{+}2{+}2$ hyper cross sparse convolution stack.}
    \label{fig:cross_block}
\end{figure}
\begin{figure}
    \centering
    \includegraphics[width=0.9\linewidth]{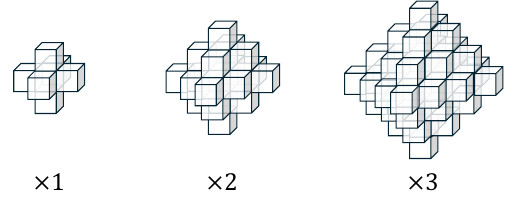}
    \captionof{figure}{Visualization of the receptive field expansion achieved by stacking hyper cross convolutions. As the number of layers increases, the effective receptive field progressively extends from axial neighbors to diagonal spatial locations.}
    \label{fig:receptive}
\end{figure}
Moreover, to further improve the computational efficiency, we adopt a hyper cross convolution kernel in both encoder and decoder. Unlike standard cubic convolution kernel, the hyper cross kernel aggregates features only from axial neighbors. For example, for an active voxel processed by a sparse convolution layer with kernel size $3$, convolution is performed with only $6$ axial neighbors instead of all 26 neighboring voxels. This design significantly reduces neighbor-matching operations and memory-access overhead, making sparse convolution more scalable and well-suited for real-time occupancy completion.

However, restricting convolution to axial neighbors inevitably limits the receptive field, particularly for diagonal or off-axis spatial interactions, which may hinder the propagation of contextual information in sparse volumes. To address this limitation, we modify the standard residual block by replacing the conventional two-layer design with a three-layer hyper cross sparse convolution stack as illustrated in Fig. \ref{fig:cross_block}. And as shown in Fig. \ref{fig:receptive}, stacking three hyper cross convolution layers enables multi-hop feature propagation along axial directions, allowing the receptive field of a single residual block to effectively cover diagonal spatial regions. This design progressively expands spatial coverage while preserving the efficiency advantage of the hyper cross kernel, achieving a favorable balance between computational cost and contextual aggregation.

\subsection{Object Contextual Representation Based Mask Decoder}
To predict the final semantic occupancy result, we design an efficient query-based mask decoder inspired by recent mask-centric transformer models. Instead of performing expensive cross-attention between learnable queries and multi-scale voxel features, as commonly adopted in previous works, we rely on a compact object contextual representation (OCR) obtained from the sparse volume produced by the completion network.

Given the final sparse voxel feature $V_0^{\prime}$, we follow the formulation of OCRNet \cite{yuan2020object} to compute class-wise object contextual representation using the proxy semantic occupancy map $P_{occ,0}$. Specifically, $P_{occ,0}$ is treated as soft region assignment weights over active voxels, enabling the aggregation of sparse voxel features into class-level region embeddings. These embeddings encode semantic group–level context and provide a descriptor of the scene structure, which is defined as
\begin{equation}
    R=\sum_{x\in V_0^{\prime}}\mathrm{Softmax}_{x}(P_{occ,0}(x))\cdot V_0^{\prime}
\end{equation}
where $R\in\mathbb{R}^{S\times C}$ denotes the object contextual representation of $V_0^{\prime}$ and $\mathrm{Softmax}_{x}(\cdot)$ denotes the softmax operation across active voxels.

In addition, we maintain a global OCR $R_{global}$ to accumulate contextual representations over the course of training. After computing the OCR embeddings from current sparse volume features, the global OCR is updated using an exponential moving average (EMA) strategy, which is defined as
\begin{equation}
    R_{global}^{\prime} = \alpha R_{global} + (1-\alpha) R 
\end{equation}
where $\alpha\in(0,1)$ denotes the momentum parameter.

Next, a set of learnable query embeddings $Q\in\mathbb{R}^{K\times C}$ interacts with the global OCR to retrieve a global contextual prior and attends to the current OCR through a cross-attention operation. By restricting attention interactions to a compact contextual space, this design substantially reduces computational overhead while retaining strong semantic reasoning capability, which is defined as
\begin{equation}
    Q^{\prime} =  \mathrm{Attn}(Q+R_{global}, R)
\end{equation}
where $K$ denotes the total number of queries and $\mathrm{Attn}(\cdot,\cdot)$ represents a standard multi-head cross-attention operator.

Afterwards, semantic scores and voxel masks are derived from the refined queries and the sparse volume features. To be specific, each query predicts semantic logits and a mask embedding through linear projections. The semantic score $P_k\in\mathbb{R}^S$ is obtained via a softmax function, while a voxel mask $M_k\in\mathbb{R}^{X_{r}\times Y_{r}\times Z_{r}}$ is predicted by computing the dot product between the mask embedding and the sparse volume features. The downsampled 3D semantic occupancy prediction is then obtained as
\begin{equation}
    Occ_{r}=\sum_{k=1}^{K}P_k\cdot M_k
\end{equation}
Finally, the full-resolution 3D semantic occupancy prediction is obtained through interpolation.

\subsection{Training Objectives}

\textbf{Query DeNoising:} To improve robustness of the model, we adopt a denoising training strategy following mask-based transformer frameworks \cite{li2022dn}. Specifically, query embeddings and their corresponding ground-truth mask targets are perturbed during training, which stabilizes optimization and improves the association between queries and voxel masks.

\textbf{Exponential Moving Average:}
To further stabilize model optimization and enhance generalization, we maintain an exponential moving average (EMA) \cite{morales2024exponential} of all model parameters. 
\begin{equation}
    \theta_{EMA} = \beta\theta_{EMA}+(1-\beta)\theta
\end{equation}
where $\beta$ denotes the momentum coefficient. This EMA mechanism operates independently of the global OCR, which instead accumulates contextual embeddings and serves as a semantic knowledge base.

\textbf{Training Loss:}
The overall training loss $\mathcal{L}_{total}$ is defined as
\begin{equation}
    \mathcal{L}_{total} = \mathcal{L}_{depth}+\mathcal{L}_{seg}+\mathcal{L}_{prune}+\mathcal{L}_{cls}+\mathcal{L}_{mask}
\end{equation}
where $\mathcal{L}_{depth}$ and $\mathcal{L}_{seg}$ denote the binary cross-entropy (BCE) loss for depth estimation and cross-entropy (CE) loss for image semantic segmentation, respectively. $\mathcal{L}_{cls}$ represents the CE loss for the semantic prediction of the refined queries. $\mathcal{L}_{mask}$ is a combination of BCE loss and Dice loss for voxel mask. $\mathcal{L}_{prune}$ corresponds to the multi-scale proxy semantic occupancy loss, which includes CE, Lovasz \cite{berman2018lovasz}, and the geometric and semantic loss introduced in \cite{Cao_2022_CVPR}.

\begin{table*}[!t]
\renewcommand\arraystretch{1.2}
\centering
\caption{Overall 3D semantic occupancy prediction result on SemanticKITTI validation set with monocular camera input. For each method, we report its venue, image backbone, parameter counts, geometry IoU, mean semantic IoU and FPS. - indicates that the corresponding information is not applicable or the official implementations are incompatible with our environment.}
\begin{tabular}{l@{\hspace{0.8cm}} c@{\hspace{0.8cm}} c@{\hspace{0.8cm}} c@{\hspace{0.8cm}}c@{\hspace{0.8cm}}c@{\hspace{0.8cm}} c@{\hspace{0.8cm}} c@{\hspace{0.8cm}}}
\toprule
\makecell*[l]{Method}&\makecell*[c]{Image Size} &\makecell*[c]{Image Backbone} &\makecell*[c]{Params.}&\makecell*[c]{IoU (val./test)}&\makecell*[c]{mIoU (val./test)}&\makecell*[c]{FPS}\\
\midrule
MonoScene\cite{Cao_2022_CVPR} & $384\times1280$ & EfficientNet-B7 & 149.6M& 37.12/34.16& 11.50/11.08 & 1.7\\
TPVFormer\cite{Huang_2023_CVPR} & $384\times1280$ & EfficientNet-B7  &140.5M& 35.61/34.25& 11.36/11.26& 2.4\\
VoxFormer-S\cite{10203337}& $384\times1280$ & ResNet-50 &58.2M&44.02/42.95&12.35/12.20& 2.6\\
OccFormer\cite{Zhang_2023_ICCV}& $384\times1280$& EfficientNet-B7 &203.4M&36.50/34.53& 13.46/12.32&4.1\\
SparseOcc\cite{tang2024sparseocc}& $384\times1280$ & EfficientNet-B7 &229.9M&36.67/-& 13.12/-&6.2\\
ProtoOcc\cite{kim2025protoocc}& $384\times1280$ & EfficientNet-B7 &167.0M&36.67/34.60& 13.89/12.74&6.4\\
\midrule
SUGOcc (Ours) & $384\times1280$ & ResNet-50 &168.5M&35.87/33.90& 14.91/13.24 & 10.2\\
\bottomrule
\end{tabular}
\label{table:overall}
\end{table*}

\begin{table*}[!t]
\renewcommand\arraystretch{1.2}
\centering
\caption{Overall 3D semantic occupancy prediction result on Occ3D-Nuscenes with single frame images input. For each method, we report its visible mask usage, image backbone, input image size, parameter counts, mean semantic IoU, RayIoU and FPS.}
\begin{tabular}{l@{\hspace{0.8cm}} c@{\hspace{0.8cm}} c@{\hspace{0.8cm}} c@{\hspace{0.8cm}}c@{\hspace{0.8cm}}c@{\hspace{0.8cm}} c@{\hspace{0.8cm}} c@{\hspace{0.8cm}}}
\toprule
\makecell*[l]{Method}&\makecell*[c]{Mask}&\makecell*[c]{Image Size} &\makecell*[c]{Image Backbone} &\makecell*[c]{Params.}&\makecell*[c]{RayIoU}&\makecell*[c]{mIoU }&\makecell*[c]{FPS}\\
\midrule
BEVDet-Occ\cite{huang2022bevdet4d}& \checkmark& $384\times704$ & ResNet-50 &35.0M& 30.81& 37.28&2.9\\
FB-Occ\cite{li2023fb}& \checkmark& $256\times704$ & ResNet-50 &68.5M& -& 37.39&10.4\\
ProtoOcc\cite{kim2025protoocc}& \checkmark& $256\times704$ & ResNet-50 &82.6M& 31.49& 39.56&10.8\\
\midrule
BEVFormer\cite{li2024bevformer}& \texttimes&$900\times1600$&ResNet-101&-&33.70&23.70&4.4\\
FB-Occ\cite{li2023fb}& \texttimes& $256\times704$ & ResNet-50 &68.5M& 35.60& 27.90&10.4\\
\midrule
SUGOcc (Ours) & \texttimes & $256\times704$ & ResNet-50 & 64.8M & 37.01 & 31.56 &  11.2\\
\bottomrule
\end{tabular}
\label{table:overall_occ3d}
\end{table*}

\section{EXPERIMENTS}

\subsection{Dataset and Metrics}

We evaluate the proposed method on two widely used datasets, SemanticKITTI \cite{behley2019iccv} and Occ3D-Nuscenes \cite{tian2024occ3d}, to validate the effectiveness of our proposed SUGOcc in both single-view and multi-view scenarios.

\begin{itemize}
    \item SemanticKITTI provides dense semantic occupancy annotations by augmenting LiDAR scans from KITTI Odometry Benchmark \cite{geiger2012we}. The prediction region is defined as a fixed 3D volume centered at the ego vehicle and the predicted region covers $51.2m$ in front of the vehicle, $25.6m$ to both left and right sides and extents from $-2.0m$ to $4.4m$ along the vertical axis. This region is discretized into a voxel grid with a resolution of $256\times256\times 32$, corresponding to a voxel size of $0.2\times0.2\times0.2m$. Each voxel is annotated with one of $20$ classes, including 19 semantic categories and 1 free space class. In this work, we focus on monocular semantic occupancy prediction, following the setting in \cite{tang2024sparseocc}.
    \item Occ3D-Nuscenes is built upon the nuScenes dataset \cite{caesar2020nuscenes} and provides large-scale 3D semantic occupancy annotations for urban driving scenes. The target volume covers $40.0m$ in front of and $40.0m$ behind the ego vehicle, $40.0m$ to both left and right sides, and extends from $-1.0m$ to $5.4m$ along the vertical axis. This region is discretized into a voxel grid with a resolution of $200\times 200\times 16$, corresponding to a voxel size of $0.4\times 0.4\times 0.4m$. Each voxel is assigned one of $18$ semantic classes, including free space and foreground object categories. Each frame in Occ3D-Nuscenes contains $6$ cameras surrounding the vehicle.
\end{itemize}


To evaluate the performance of semantic occupancy prediction, we employ the geometric Intersection over Union (IoU), per class semantic IoU and mean IoU over all semantic categories as the primary metrics. In addition, to evaluate the computational efficiency, we measure the inference speed in terms of frames per second (FPS), which reflects the practical efficiency and real-time capability of the model.

\subsection{Implementation Details}
For each frame, the input image is first cropped into a resolution of $384\times1280$ for SemanticKITTI and $256\times704$ for Occ3D-Nuscenes. Then, we adopt the image encoder with ResNet-50 backbone from MaskDINO \cite{li2023mask} and BEVDet \cite{huang2022bevdet4d}, respectively, as the pretrained image feature extractors for SemanticKITTI and Occ3D-Nuscenes, respectively, producing multi-scale image features. The multi-scale image features are then fused into a unified representation at $\frac{1}{16}$ resolution using SECONDFPN \cite{sindagi2019mvx} and FPN \cite{lin2017feature}. The fused image feature is subsequently fed into depth and semantic predictors to obtain 2D depth distribution and semantic segmentation. Guided by the depth and semantic priors, the fused image feature is lifted into the 3D space through view transformation, producing a sparse volume feature at $\frac{1}{2}$ resolution for SemanticKITTI and $1\times$ resolution for Occ3D-Nuscenes of the target voxel grid. For the cascade sparse completion network, we employ two hyper cross residual blocks in both the encoder and decoder. In the OCR-based mask decoder, the number of learnable queries is set to $100$, and one cross-attention layer followed by one self-attention layer is applied. During voxel mask prediction, a learnable embedding is used to fill empty positions in the sparse volume feature, following the practice in \cite{tang2024sparseocc}.

During training, standard data augmentation strategies for depth estimation, semantic segmentation, and semantic occupancy prediction are applied. In addition, we employ a query denoising strategy with $20$ denoising queries, together with an EMA scheme for model optimization. The model is trained using the AdamW optimizer with a learning rate of $4\times 10^{-4}$ and a total batch size of $4$ in $20$ epochs for SemanticKITTI and a learning rate of $4\times 10^{-4}$ and a total batch size of $16$ in $30$ epochs for Occ3D-Nuscenes on $8$ NVIDIA H100. For efficiency evaluation, all models are measured on an NVIDIA RTX 4090 using PyTorch fp32 backend with a batch size of $1$. The FPS measurement follows the protocol in \cite{chen2025alocc}.

\begin{table*}[htbp]
\renewcommand\arraystretch{1.2}
\setlength{\tabcolsep}{3pt}
\centering
\caption{3D semantic occupancy prediction result of each semantic category on the SemanticKITTI validation set with a monocular camera. The bold numbers indicate the best results.}
\begin{center}
\begin{tabular}{l | c |c| c@{\hspace{0.15cm}} c@{\hspace{0.15cm}} c@{\hspace{0.15cm}} c@{\hspace{0.15cm}} c@{\hspace{0.15cm}} 
c@{\hspace{0.15cm}} c@{\hspace{0.15cm}} c@{\hspace{0.15cm}}
c@{\hspace{0.15cm}} c@{\hspace{0.15cm}} c@{\hspace{0.15cm}} c@{\hspace{0.15cm}} c@{\hspace{0.15cm}}c@{\hspace{0.15cm}}c@{\hspace{0.15cm}}c@{\hspace{0.15cm}}c@{\hspace{0.15cm}}c@{\hspace{0.15cm}}c@{\hspace{0.15cm}}}
\toprule
\multicolumn{1}{l|}{Method}
&\multicolumn{1}{c|}{\makecell*[b]{IoU}}
& \multicolumn{1}{c|}{mIoU}
& \makecell*[b]{\rotatebox{90}{\raisebox{1\height}{\colorbox{kitti_car}{}}\,Car}}
& \makecell*[b]{\rotatebox{90}{\raisebox{1\height}{\colorbox{kitti_bicycle}{}}\,Bicycle}}
& \makecell*[b]{\rotatebox{90}{\raisebox{1\height}{\colorbox{kitti_motorcycle}{}}\,Motorcycle}}
& \makecell*[b]{\rotatebox{90}{\raisebox{1\height}{\colorbox{kitti_truck}{}}\,Truck}}
& \makecell*[b]{\rotatebox{90}{\raisebox{1\height}{\colorbox{kitti_other-vehicle}{}}\,Other-Vehicle}}
& \makecell*[b]{\rotatebox{90}{\raisebox{1\height}{\colorbox{kitti_person}{}}Person}}
& \makecell*[b]{\rotatebox{90}{\raisebox{1\height}{\colorbox{kitti_bicyclist}{}}Bicyclist}}
& \makecell*[b]{\rotatebox{90}{\raisebox{1\height}{\colorbox{kitti_motorcyclist}{}}Motorcyclist}}
& \makecell*[b]{\rotatebox{90}{\raisebox{1\height}{\colorbox{kitti_road}{}}\,Road}}
& \makecell*[b]{\rotatebox{90}{\raisebox{1\height}{\colorbox{kitti_parking}{}}\,Parking}}
& \makecell*[b]{\rotatebox{90}{\raisebox{1\height}{\colorbox{kitti_sidewalk}{}}\,Sidewalk}}
& \makecell*[b]{\rotatebox{90}{\raisebox{1\height}{\colorbox{kitti_other-ground}{}}\,Other-Ground}}
& \makecell*[b]{\rotatebox{90}{\raisebox{1\height}{\colorbox{kitti_building}{}}\,Building}}
& \makecell*[b]{\rotatebox{90}{\raisebox{1\height}{\colorbox{kitti_fence}{}}\,Fence}}
& \makecell*[b]{\rotatebox{90}{\raisebox{1\height}{\colorbox{kitti_vegetation}{}}\,Vegetation}}
& \makecell*[b]{\rotatebox{90}{\raisebox{1\height}{\colorbox{kitti_trunk}{}}\,Trunk}}
& \makecell*[b]{\rotatebox{90}{\raisebox{1\height}{\colorbox{kitti_terrain}{}}Terrain}}
& \makecell*[b]{\rotatebox{90}{\raisebox{1\height}{\colorbox{kitti_pole}{}}\,Pole}}
& \makecell*[b]{\rotatebox{90}{\raisebox{1\height}{\colorbox{kitti_trafficsign}{}}\,Trafficsign}}\\
\midrule
MonoScene\cite{Cao_2022_CVPR} & 37.12 &11.50& 23.55& 0.20& 0.77& 7.83& 3.59& 1.79& 1.03& 0.00& 57.47& 15.72& 27.05& 0.87& 14.24& 6.39& 18.12& 2.57& 30.76& 4.11& 2.48\\
TPVFormer\cite{Huang_2023_CVPR}& 35.61& 11.36& 23.81& 0.36& 0.05& 8.08& 4.35& 0.51& 0.89& \textbf{0.00}& 56.50& 20.60& 25.87& 0.85& 13.88& 5.94& 16.92& 2.26& 30.38& 3.14& 1.52\\
VoxFormer-S\cite{10203337} &\textbf{44.02}&12.35&\textbf{25.79}&0.59&0.51&5.63&3.77&1.78&3.32&0.00&54.76&15.50&26.35&0.70&\textbf{17.65}&\textbf{7.64}&\textbf{24.39}&\textbf{5.08}&29.96&\textbf{7.11}&\textbf{4.18} \\
OccFormer\cite{Zhang_2023_ICCV}& 36.67&13.46&25.09& 0.81& 1.19& \textbf{25.53}& 8.52& 2.78& 2.82& 0.00& 58.85& 19.61& 26.88& 0.31& 14.40& 5.61& 19.63& 3.93& 33.62& 4.26& 2.86\\
SparseOcc\cite{tang2024sparseocc}& 36.50& 13.12& 25.09&0.78&0.89&18.07&8.94&3.68&0.62&0.00&\textbf{59.59}&20.44&\textbf{29.68}&0.47&14.40&6.73&18.89&3.46&31.06&3.89&2.60\\
ProtoOcc\cite{kim2025protoocc}& 36.67&13.89& 24.88& 3.65& 2.06& 20.65& 13.01& 4.50& 0.50& 0.00& 58.84& 20.61& 28.34& 0.82& 16.46& 6.72& 19.55& 4.33& 32.43& 4.49& 1.95\\
\midrule
SUGOcc (Ours) & 
35.87&\textbf{14.91}& 24.28&\textbf{4.15}&\textbf{6.38}&23.06&\textbf{16.96}&\textbf{5.27}&\textbf{5.56}&0.00&56.20&\textbf{21.89}&28.23&\textbf{2.50}&15.24&7.55&20.00&4.03&\textbf{33.76}&4.99&3.43\\
\bottomrule
\end{tabular}
\end{center}
\label{table:kitti}
\end{table*}

\begin{table*}[htbp]
\renewcommand\arraystretch{1.2}
\setlength{\tabcolsep}{3pt}
\centering
\caption{3D semantic occupancy prediction result of each semantic category on the SemanticKITTI hidden test set with monocular camera input.}
\begin{center}
\begin{tabular}{l | c |c| c@{\hspace{0.15cm}} c@{\hspace{0.15cm}} c@{\hspace{0.15cm}} c@{\hspace{0.15cm}} c@{\hspace{0.15cm}} 
c@{\hspace{0.15cm}} c@{\hspace{0.15cm}} c@{\hspace{0.15cm}}
c@{\hspace{0.15cm}} c@{\hspace{0.15cm}} c@{\hspace{0.15cm}} c@{\hspace{0.15cm}} c@{\hspace{0.15cm}}c@{\hspace{0.15cm}}c@{\hspace{0.15cm}}c@{\hspace{0.15cm}}c@{\hspace{0.15cm}}c@{\hspace{0.15cm}}c@{\hspace{0.15cm}}}
\toprule
\multicolumn{1}{l|}{Method}
&\multicolumn{1}{c|}{\makecell*[b]{IoU}}
& \multicolumn{1}{c|}{mIoU}
& \makecell*[b]{\rotatebox{90}{\raisebox{1\height}{\colorbox{kitti_car}{}}\,Car}}
& \makecell*[b]{\rotatebox{90}{\raisebox{1\height}{\colorbox{kitti_bicycle}{}}\,Bicycle}}
& \makecell*[b]{\rotatebox{90}{\raisebox{1\height}{\colorbox{kitti_motorcycle}{}}\,Motorcycle}}
& \makecell*[b]{\rotatebox{90}{\raisebox{1\height}{\colorbox{kitti_truck}{}}\,Truck}}
& \makecell*[b]{\rotatebox{90}{\raisebox{1\height}{\colorbox{kitti_other-vehicle}{}}\,Other-Vehicle}}
& \makecell*[b]{\rotatebox{90}{\raisebox{1\height}{\colorbox{kitti_person}{}}Person}}
& \makecell*[b]{\rotatebox{90}{\raisebox{1\height}{\colorbox{kitti_bicyclist}{}}Bicyclist}}
& \makecell*[b]{\rotatebox{90}{\raisebox{1\height}{\colorbox{kitti_motorcyclist}{}}Motorcyclist}}
& \makecell*[b]{\rotatebox{90}{\raisebox{1\height}{\colorbox{kitti_road}{}}\,Road}}
& \makecell*[b]{\rotatebox{90}{\raisebox{1\height}{\colorbox{kitti_parking}{}}\,Parking}}
& \makecell*[b]{\rotatebox{90}{\raisebox{1\height}{\colorbox{kitti_sidewalk}{}}\,Sidewalk}}
& \makecell*[b]{\rotatebox{90}{\raisebox{1\height}{\colorbox{kitti_other-ground}{}}\,Other-Ground}}
& \makecell*[b]{\rotatebox{90}{\raisebox{1\height}{\colorbox{kitti_building}{}}\,Building}}
& \makecell*[b]{\rotatebox{90}{\raisebox{1\height}{\colorbox{kitti_fence}{}}\,Fence}}
& \makecell*[b]{\rotatebox{90}{\raisebox{1\height}{\colorbox{kitti_vegetation}{}}\,Vegetation}}
& \makecell*[b]{\rotatebox{90}{\raisebox{1\height}{\colorbox{kitti_trunk}{}}\,Trunk}}
& \makecell*[b]{\rotatebox{90}{\raisebox{1\height}{\colorbox{kitti_terrain}{}}Terrain}}
& \makecell*[b]{\rotatebox{90}{\raisebox{1\height}{\colorbox{kitti_pole}{}}\,Pole}}
& \makecell*[b]{\rotatebox{90}{\raisebox{1\height}{\colorbox{kitti_trafficsign}{}}\,Trafficsign}}\\
\midrule
MonoScene\cite{Cao_2022_CVPR} & 34.16 &11.08& 18.80& 0.50& 0.70& 3.30& 4.40& 1.00& 1.40& 0.40& 54.70& 24.80& 27.10& 5.70& 14.40& 11.10& 14.90& 2.40& 19.50& 3.30& 2.10\\
TPVFormer\cite{Huang_2023_CVPR}& 34.25& 11.26& 19.20& 1.00& 0.50& 3.70& 2.30& 1.10& 2.40& 0.30& 55.10& 27.40& 27.20& 6.50& 14.80& 11.00& 13.90& 2.60& 20.40& 2.90& 1.50\\
VoxFormer-S\cite{10203337} 
&\textbf{42.95}&12.20&20.80&1.00&0.70&3.50&3.70&1.40&2.60&0.20&53.90&21.10&25.30&5.60&\textbf{19.80}&11.10&\textbf{22.40}&\textbf{7.50}&21.30&\textbf{5.10}&\textbf{4.90} \\
OccFormer\cite{Zhang_2023_ICCV}& 34.53&12.32&21.60& 1.50& 1.70& 1.20& 3.20& 2.20& 1.10& 0.20& \textbf{55.90}& \textbf{31.50}& \textbf{30.30}& 6.50& 15.70& 11.90& 16.80& 3.90& 21.30& 3.80& 3.70\\
ProtoOcc\cite{kim2025protoocc}& 34.60&12.74&\textbf{21.70} & 3.20 & 1.70 & 2.60 & 4.90 & 3.90 & 1.20 & 0.00 & 55.50 & 30.00 & 28.80 & \textbf{10.60} & 17.00 & 12.70 & 16.60 & 4.10 & 22.00 & 4.10 & 1.50 \\
\midrule
SUGOcc (Ours) & 
33.90&\textbf{13.24}&  21.40 & \textbf{4.30} & \textbf{3.90} & \textbf{4.40} & \textbf{5.10} & \textbf{5.00} & \textbf{4.70} & \textbf{1.00} & 52.00 & 29.90 & 29.20 & 10.30 & 16.60 & \textbf{12.90} & 17.00 & 4.20 & \textbf{22.40} & 4.00 & 4.20 \\
\bottomrule
\end{tabular}
\end{center}
\label{table:kitti_test}
\end{table*}

\subsection{Main Results}
We compare the proposed method with several representative and state-of-the-art 3D semantic occupancy prediction methods on the SemanticKITTI dataset, including MonoScene \cite{Cao_2022_CVPR}, TPVFormer \cite{Huang_2023_CVPR}, VoxFormer \cite{10203337}, OccFormer \cite{Zhang_2023_ICCV}, SparseOcc \cite{tang2024sparseocc} and ProtoOcc \cite{kim2025protoocc}, as well as several earlier methods such as LMSCNet \cite{9320442}, 3DSketch \cite{Cao_2022_CVPR}, AICNet \cite{chen20203d}, and JS3C-Net\cite{yan2021sparse}, whose results are reproduced and reported by MonoScene. For performance comparison, the reported results of these methods are derived directly from their original papers. To ensure a fair evaluation of efficiency, we re-evaluate all models under their official configurations in a unified environment. Models with publicly available checkpoints are evaluated through direct inference. For methods without released weights, we retrain them from scratch following their reported configurations.

\begin{table*}[htbp]
\renewcommand\arraystretch{1.2}
\setlength{\tabcolsep}{3pt}
\centering
\caption{Detail RayIoU result of each semantic category on the Occ3D-Nuscenes. The bold numbers indicate the best results.}
\begin{center}
\begin{tabular}{l | c |c| c@{\hspace{0.15cm}} c@{\hspace{0.15cm}} c@{\hspace{0.15cm}} c@{\hspace{0.15cm}} 
c@{\hspace{0.15cm}} c@{\hspace{0.15cm}} c@{\hspace{0.15cm}}
c@{\hspace{0.15cm}} c@{\hspace{0.15cm}} c@{\hspace{0.15cm}} c@{\hspace{0.15cm}} c@{\hspace{0.15cm}}c@{\hspace{0.15cm}}c@{\hspace{0.15cm}}c@{\hspace{0.15cm}}c@{\hspace{0.15cm}}c@{\hspace{0.15cm}}}
\toprule
\multicolumn{1}{l|}{Method}
&\multicolumn{1}{c|}{\makecell*[b]{Mask}}
&\multicolumn{1}{c|}{\makecell*[b]{RayIoU}}
& \makecell*[b]{\rotatebox{90}{\raisebox{1\height}{\colorbox{nus_others}{}}\,Others}}
& \makecell*[b]{\rotatebox{90}{\raisebox{1\height}{\colorbox{nus_barrier}{}}\,Barrier}}
& \makecell*[b]{\rotatebox{90}{\raisebox{1\height}{\colorbox{nus_bicycle}{}}\,Bicycle}}
& \makecell*[b]{\rotatebox{90}{\raisebox{1\height}{\colorbox{nus_bus}{}}\,Bus}}
& \makecell*[b]{\rotatebox{90}{\raisebox{1\height}{\colorbox{nus_car}{}}Car}}
& \makecell*[b]{\rotatebox{90}{\raisebox{1\height}{\colorbox{nus_construction_vehicle}{}}\,Const. Veh.}}
& \makecell*[b]{\rotatebox{90}{\raisebox{1\height}{\colorbox{nus_motorcycle}{}}Motorcycle}}
& \makecell*[b]{\rotatebox{90}{\raisebox{1\height}{\colorbox{nus_pedestrian}{}}Pedestrian}}
& \makecell*[b]{\rotatebox{90}{\raisebox{1\height}{\colorbox{nus_traffic_cone}{}}Traffic Cone}}
& \makecell*[b]{\rotatebox{90}{\raisebox{1\height}{\colorbox{nus_trailer}{}}\,Trailer}}
& \makecell*[b]{\rotatebox{90}{\raisebox{1\height}{\colorbox{nus_truck}{}}\,Trunk}}
& \makecell*[b]{\rotatebox{90}{\raisebox{1\height}{\colorbox{nus_driveable_surface}{}}\,Drive. Surf.}}
& \makecell*[b]{\rotatebox{90}{\raisebox{1\height}{\colorbox{nus_other_flat}{}}\,Other Flat}}
& \makecell*[b]{\rotatebox{90}{\raisebox{1\height}{\colorbox{nus_sidewalk}{}}\,Sidewalk}}
& \makecell*[b]{\rotatebox{90}{\raisebox{1\height}{\colorbox{nus_terrain}{}}\,Terrain}}
& \makecell*[b]{\rotatebox{90}{\raisebox{1\height}{\colorbox{nus_manmade}{}}\,Manmade}}
& \makecell*[b]{\rotatebox{90}{\raisebox{1\height}{\colorbox{nus_vegetation}{}}\,Vegetation}}\\
\midrule
BEVDet-Occ\cite{huang2022bevdet4d} & \checkmark & 30.81&6.90&42.30&18.43&55.36&54.60&\textbf{26.56}&18.80&27.53
&22.03&30.20&48.30&41.90&19.60&20.23&16.90&41.43&32.76\\

ProtoOcc\cite{kim2025protoocc}& \checkmark&31.49&4.20&42.56&24.93&58.26&56.26&24.83&27.10&32.66&28.23&29.00&49.33&40.13&19.63&20.46&17.30&35.46&24.93\\
\midrule
BEVFormer\cite{li2024bevformer}& \texttimes&33.70 &5.00 & 42.20& 18.20& 55.20& \textbf{57.10}& 22.70& 21.30& 31.00& 27.10& \textbf{30.70}& 49.40& 58.40& 30.40& 29.40& 31.70& 36.30 &26.50\\

FB-Occ\cite{li2023fb}&\texttimes&35.60&\textbf{10.50}&\textbf{44.80}&25.60&55.60&51.70&22.60&\textbf{27.20}&\textbf{34.30}&30.30&23.70&44.10&65.50&33.30&31.40&32.50&\textbf{39.60}&\textbf{33.30}\\
\midrule
SUGOcc (Ours) & \texttimes & \textbf{37.01} & 10.30 &42.90&\textbf{25.80} &\textbf{64.03}  &55.80&24.67&25.87&34.20&\textbf{30.67}&26.87&\textbf{49.50}&\textbf{66.40}&\textbf{35.27}&32.37&33.00&\textbf{39.60}&32.00 \\
\bottomrule
\end{tabular}
\end{center}
\label{table:occ3d_detail}
\end{table*}
\begin{table}[htbp]
\renewcommand\arraystretch{1.2}
\centering
\caption{Ablation study for the impact of each proposed component on SemanticKITTI validation set.}
\begin{center}
\begin{tabular}{l | c |c|c}
\toprule
\multicolumn{1}{c|}{Condition} &\multicolumn{1}{c|}{IoU} &\multicolumn{1}{c|}{mIoU} &\multicolumn{1}{c}{FPS}\\
\midrule
Baseline & 35.87& 14.91 & 10.2 \\
w/o Semantics and Uncertainty Guided LSS & 35.67 & 14.38& 9.7\\
w/o Hyper-Cross Kernel &35.12& 14.36& 7.5 \\
w/o Object Contextual Representation  &35.21& 14.25& 10.4 \\
\bottomrule
\end{tabular}
\end{center}
\label{table:modules}
\end{table}

\begin{table}[htbp]
\renewcommand\arraystretch{1.2}
\centering
\caption{Ablation study for the impact of each training Strategy on SemanticKITTI validation set.}
\begin{center}
\begin{tabular}{l | c |c}
\toprule
\multicolumn{1}{c|}{Condition} &\multicolumn{1}{c|}{IoU}&\multicolumn{1}{c}{mIoU} \\
\midrule
Baseline &35.87 & 14.91\\
w/o Query Denoising & 35.67& 14.22\\
w/o Exponential Moving Average  &35.98& 13.34\\
\bottomrule
\end{tabular}
\end{center}
\label{table:training}
\end{table}

Table \ref{table:overall} presents an overall comparison on the SemanticKITTI validation set, including the image backbone, number of parameters, geometric IoU, mean IoU (mIoU) over all semantic classes and inference speed measured in FPS. As shown in the table, although using a relatively lightweight ResNet-50 based image backbone, our method achieves the highest mIoU and FPS among all compared approaches. In particular, compared with the state-of-the-art method ProtoOcc, the proposed method improves mIoU by $7.34\%$ while achieving a $59.38\%$ increase in FPS. Note that the parameter counts and FPS of MonoScene, TPVFormer and VoxFormer are not reported as their official implementations rely on legacy versions of PyTorch and CUDA that are incompatible with our evaluation environment, while the image backbones of those earlier methods are not applicable, which are both marked as shown in the table.

Table \ref{table:overall_occ3d} reports an overall comparison on the Occ3D-Nuscenes, including whether the models are trained with visible mask, their image backbone, input image size, number of parameters, RayIoU, mIoU and FPS. To be specific, due to the visible mask provided by Occ3D-Nuscenes labels more than $80\%$ of the voxels as 255 (unknown), which conflicts with our pruning model design, we train our model without visible mask. It is observed that, models trained with visible mask generally tend to achieve higher mIoU, since excessive unknown labels lead to overestimation of object surfaces. Therefore, SparseOcc \cite{liu2024fully} introduces RayIoU to precisely measure the performance around the surface of each object. As shown in the table, our proposed method achieves the highest RayIoU while also attaining the best FPS. This demonstrates that our method provides more accurate surface-level predictions, while maintaining a favorable accuracy–efficiency trade-off.

Table \ref{table:kitti} reports the per class IoU results on the SemanticKITTI validation set, providing a fine-grained comparison of semantic occupancy prediction performance across different object and scenes. A closer inspection reveals that the overall improvement is primarily driven by consistent gains on small, sparse, and dynamic object categories, which are traditionally challenging for vision-based semantic occupancy prediction. In particular, the proposed method achieves higher IoU scores on Bicycle ($4.15$), Motorcycle ($6.38$), Other-Vehicle ($16.96$), Person ($5.27$), and Bicyclist ($5.56$) when compared with strong recent baselines such as ProtoOcc. These categories typically occupy limited spatial extents and exhibit complex geometric structures, and improvements on them contribute significantly to the overall mean IoU. Overall, the IoU results of each semantic category demonstrate that the proposed method enhances semantic occupancy prediction particularly on geometrically challenging categories, while maintaining competitive performance on other classes, resulting in a higher mIoU.

Table \ref{table:occ3d_detail} presents the per class RayIoU results on Occ3D-Nuscenes dataset. As we can see from the table, our proposed method achieves consistently competitive performance across all semantic categories. Notably, it demonstrates significant improvements on several challenging categories, including Bus ($64.03$), Traffic Cone ($30.67$), and Other Flat ($35.27$), which outperforms previous methods by clear margins. These gains demonstrate the effectiveness of our proposed framework in capturing both large-scale structured objects and small or ambiguous categories.

\subsection{Ablation Study}
In this section, we conduct ablation studies to analyze the contribution of individual components in the proposed framework, as well as the effects of key pruning hyperparameters and training strategies.

\begin{table}[t]
\renewcommand\arraystretch{1.2}
\centering
\caption{Ablation study for the impact of the three pruning ratios during the prediction pipeline on SemanticKITTI validation set. Lifting voxels represent the number of features obtained from view transformation.}
\begin{center}
\begin{tabular}{l | c |c |c|c }
\toprule
\multicolumn{1}{c|}{($\tau_s\text{, }\tau_d\text{, }\tau_p$)} &\multicolumn{1}{c|}{IoU}&\multicolumn{1}{c|}{mIoU} &\multicolumn{1}{c|}{Lifting Voxels}&\multicolumn{1}{c}{FPS}\\
\midrule
$(0.0\text{, }0.0\text{, }0.0)$ &35.84& 14.50 &\multirow{2}{*}{59238}& 7.8\\
$(0.0\text{, }0.0\text{, }0.1)$ & 35.59& 14.71&&9.3\\
\midrule
$(0.1\text{, }0.1\text{, }0.0)$ &36.71& 15.10&\multirow{3}{*}{45831}&8.2\\
$(0.1\text{, }0.1\text{, }0.1)$ &35.87 & 14.91 && 10.2\\
$(0.1\text{, }0.1\text{, }0.3)$ &35.83& 14.50&&10.5\\
\midrule
$(0.3\text{, }0.3\text{, }0.1)$ &35.81& 14.54&\multirow{2}{*}{40967}&10.4\\
$(0.3\text{, }0.3\text{, }0.3)$ & 36.35& 14.52&&10.8\\
\midrule
$(0.5\text{, }0.5\text{, }0.5)$ & 35.85& 13.76&36867& 11.3\\
\bottomrule
\end{tabular}
\end{center}
\label{table:pruning}
\end{table}

\begin{table}[t]
\renewcommand\arraystretch{1.2}
\centering
\caption{Ablation study for the impact of semantic and depth priors for pruning process during view transformation on SemanticKITTI validation set.}
\begin{center}
\begin{tabular}{c | c| c |c |c|c }
\toprule
\multicolumn{1}{c|}{Sem.} &\multicolumn{1}{c|}{Depth}&\multicolumn{1}{c|}{IoU}&\multicolumn{1}{c|}{mIoU} &\multicolumn{1}{c|}{Lifting Voxels}&\multicolumn{1}{c}{FPS}\\
\midrule
& &35.84& 14.71 &59238& 9.3\\
\checkmark& &35.81&14.84&52067&9.4\\
& \checkmark& 36.28 &14.40&51560&9.6\\
\checkmark& \checkmark& 35.87 & 14.91 &45831& 10.2\\
\bottomrule
\end{tabular}
\end{center}
\label{table:prior}
\end{table}

\begin{figure*}[htbp]
    \subfloat[Camera Views]{\includegraphics[width=0.99\linewidth]{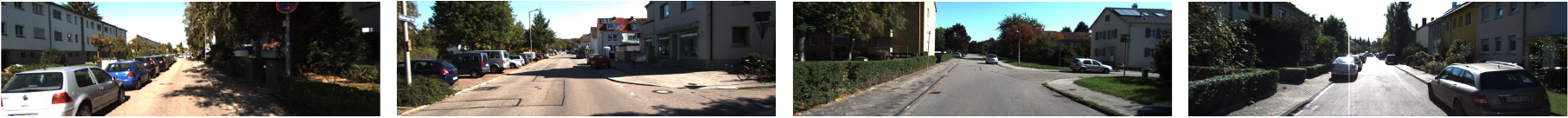}}
    \hfill
    \vspace{3pt}
    \subfloat[Downsampled Semantic Segmentation]{\includegraphics[width=0.99\linewidth]{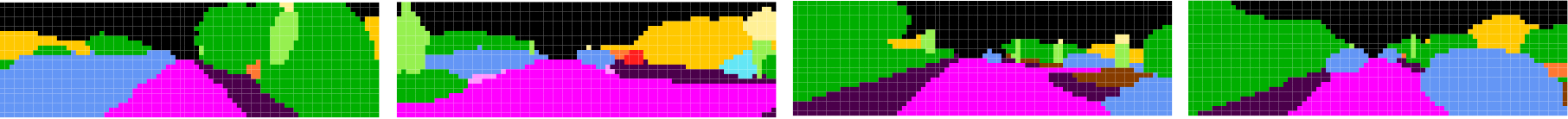}}
    \hfill
    \vspace{3pt}
    \subfloat[Downsampled Depth Distribution]{\includegraphics[width=0.99\linewidth]{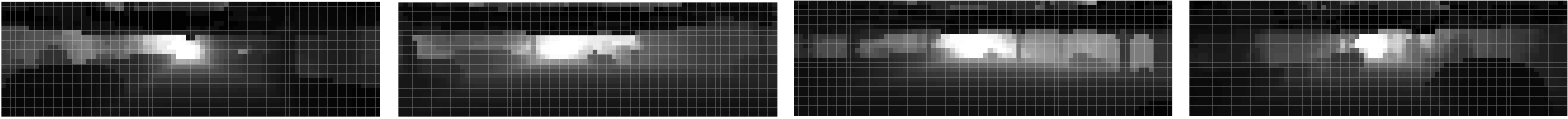}}
    \hfill
    \vspace{3pt}
    \subfloat[Guided Sparse View Projection]{\includegraphics[width=0.99\linewidth]{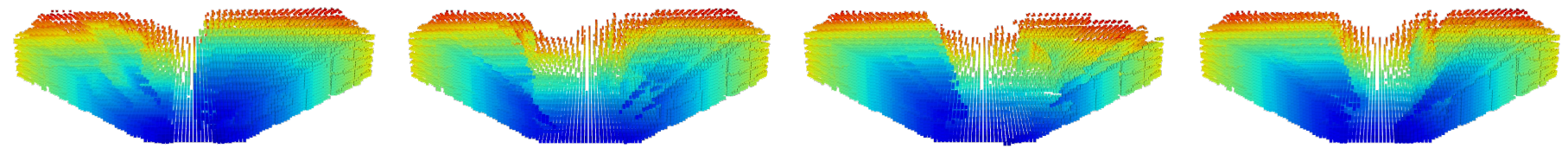}}
    \caption{Qualitative visualization of the explicit semantics and uncertainty guided LSS process. From top to bottom: (a) camera views, (b) downsampled semantic segmentation, (c) downsampled depth distribution, and (d) the resulting sparse view projections guided by semantic and depth uncertainty.}
    \label{fig:projection}
\end{figure*}

Table \ref{table:modules} presents an ablation study that analyzes the contribution of each component in the proposed methods. Starting from the full model as the baseline, we remove one component at a time and report the corresponding changes in geometric IoU, mIoU, and FPS. Removing the explicit distance encoding results in a noticeable decrease in mIoU by $0.53$, indicating that explicit distance modeling plays an important role in preserving geometric structure during view transformation. Replacing the hyper cross convolution kernel with a standard cubic kernel leads to a clear reduction in FPS by $2.6$, together with a decrease in mIoU by $0.55$. This highlights the effectiveness of hyper cross sparse convolution in improving computational efficiency while maintaining representation quality. Excluding the object contextual representation module causes a significant drop in mIoU by $0.66$, while FPS increase slightly by $0.3$. This suggests that object context representation is beneficial for refining voxel-wise mask predictions while introducing only a small computational overhead. Overall, the results demonstrate that each proposed component contributes positively to the final performance, and their combination yields the best balance between accuracy and efficiency.

Table \ref{table:training} reports the ablation study on the impact of the two training strategies employed in our framework, query denoising and exponential moving average. Removing the query denoising strategy results in a noticeable decrease in mIoU by $0.69$, indicating that denoising queries facilitate more stable query learning and improved semantic discrimination during training. In parallel, excluding the exponential moving average strategy leads to a more pronounced performance reduction in mIoU by $1.57$, suggesting that EMA plays a critical role in stabilizing model optimization and improving generalization. Overall, both query denoising and EMA contribute positively to the final performance, with EMA showing a stronger impact on model optimization stability and generalization ability.

Table \ref{table:pruning} presents the impact of different pruning thresholds settings on performance and efficiency of semantic occupancy prediction. The pruning strategy jointly considers three thresholds, $\tau_d$, $\tau_s$ and $\tau_p$, which correspond to semantic non-empty filtering at image level, depth uncertainty filtering based on the prefix sum of depth distributions, and empty voxel pruning during generative upsampling, respectively. As shown in Table \ref{table:pruning}, the non-pruning setting $(0,0,0)$ yields the lowest performance with a mIoU of $14.50$ and a lowest FPS of $7.8$. This confirms that lifting and decoding all positions introduces substantial redundancy and noise. Once pruning is enabled, both accuracy and speed exhibit clear and consistent improvements. Increasing the semantic and uncertainty thresholds $(\tau_s, \tau_d)$ from $(0,0)$ to $(0.1,0.1)$ improves geometric IoU from $35.84$ to $36.71$ and mIoU from $14.50$ to $15.10$, together with a slight increase in FPS from $7.9$ to $8.2$. This improvement is accompanied by a reduction in the number of the active voxels from $59238$ to $45831$, indicating that filtering out image positions with a high confidence of being empty during view transformation leads to a cleaner and more compact sparse 3D representation. Furthermore, when enabling empty voxel pruning during generative upsampling by increasing $\tau_p$ from $0.0$ to $0.1$ while disabling image level pruning, FPS increases from $7.8$ to $9.3$ and mIoU slightly improves from $14.50$ to $14.71$. This suggests that active voxel control during the cascade generative upsampling stage not only improves computational efficiency but is also beneficial to performance. Combining the image level pruning with the upsampling stage pruning yields the most favorable trade-off between accuracy and efficiency, achieving an mIoU of $14.91$ and a FPS of $10.2$. When the pruning thresholds are further increased, the number of active voxels decreases only marginally, while both geometric IoU and mIoU begin to decline, indicating that overly aggressive pruning negatively affects semantic completeness.

\begin{figure*}[htbp]
    \includegraphics[width=0.99\linewidth]{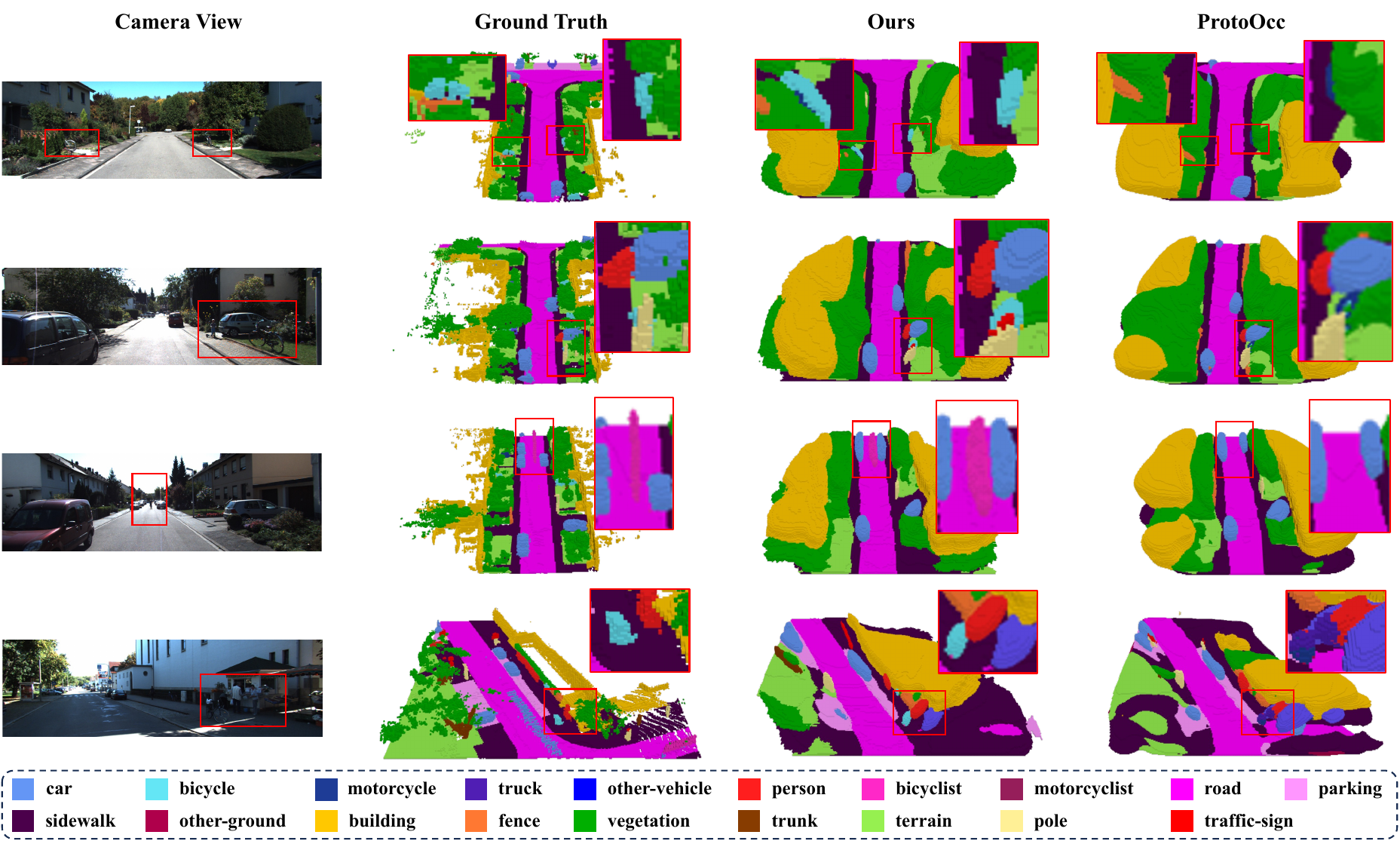}
    \caption{Qualitative visualization results of our proposed SUGOcc and the compared method ProtoOcc, along with the camera view and ground truth annotations. The zoom in region highlights the key region of interest.}
    \label{fig:visual}
\end{figure*}

Table \ref{table:prior} presents an ablation study on the impact of semantic and depth priors that used for pruning during view transformation procedure. The pruning ratio here is set to be $0.1$. When only semantic priors are applied, the model achieves a slight improvement in mIoU, indicating that semantic cues are effective in filtering out empty regions. Although the number of lifted voxels is reduced, the FPS remains nearly unchanged. In contrast, using only depth priors leads to a moderate increase in FPS due to more aggressive pruning, but results in a degradation in mIoU. This is mainly because depth priors are less reliable in empty or ambiguous regions, which may cause incorrect voxel removal. When both semantic and depth priors are jointly utilized, the model achieves the best performance in terms of both mIoU and FPS. This result highlights the complementarity of semantic and depth priors in balancing accuracy and efficiency.

\subsection{Qualitative Visualization}
Fig. \ref{fig:projection} presents the qualitative visualizations of the explicit semantics and uncertainty guided LSS process. Given the input images in Fig. \ref{fig:projection}(a), the corresponding predictions of downsampled semantic segmentation and depth distributions are shown in Fig. \ref{fig:projection}(b) and (c), respectively. Based on these cues, the sparse view projection results are illustrated in Fig. \ref{fig:projection}(d), where different colors indicate the distance to camera plane. It can be observed that the proposed method effectively suppresses projections in empty regions, while preserving structurally meaningful areas. By jointly leveraging semantic information and depth uncertainty, the proposed method produces a more compact and informative projection space, which benefits subsequent processing.

Fig. \ref{fig:visual} presents qualitative visualizations of the proposed method and the compared method ProtoOcc on the SemanticKITTI validation set. We select representative scenes and visualize the predicted 3D semantic occupancy results alongside the corresponding camera views and ground truth annotations. Overall, our method produces more complete and structurally consistent semantic occupancy predictions, particularly for small, dynamic, and thin structured objects.

As illustrated in the zoomed regions, our approach shows clear advantages in predicting classes like persons, bicycle, and motorcycle. In contrast, the compared method exhibit object missing or semantic confusion to varying degrees across different scenarios. For example, in the second scene, although a person, a bicycle, and a car are present on the right side of the camera view, the compared method incorrectly predicts the bicycle. In the third scene, the compared method fail to detect a bicycle located farther away in the middle of the road. Such errors could pose potential safety risks to autonomous driving system. In comparison, our proposed method handle these challenging scenes more effectively by consistently predicting clearer object extents and more accurate semantic labels. 

\section{Conclusion and Discussion}
This work introduces SUGOcc, an efficient 3D semantic occupancy prediction network that utilizes explicit semantics and uncertainty guided sparse learning with explicit distance encoding. By integrating semantic and depth priors to selectively lift informative image projections during view transformation, the proposed method constructs a robust and compact sparse 3D representation, effectively reducing redundancy while preserving geometric and semantic consistency. Built upon this sparse representation, a lightweight cascade sparse completion module efficiently reconstructs scene geometry and semantics via hyper cross sparse convolution, generative upsampling and adaptive pruning. To further refine semantic occupancy prediction, an object contextual representation based mask decoder aggregates global semantic context from sparse volume features and performs lightweight query-context interactions, avoiding expensive dense voxel wise attention while maintaining strong semantic reasoning capability. Extensive experiments on SemanticKITTI and Occ3D-Nuscenes datasets demonstrates that the proposed method achieves state-of-the-art performance while significantly improving inference efficiency, enabling real-time deployment in autonomous driving system.

The primary limitation of this work lies in its reliance on single-frame camera inputs, without leveraging temporal information or multi-vehicle collaborative cues. Future work could integrate temporal modeling, multi-modal sensory inputs, and collaborative perception among multiple agents to further enhance the performance and scalability of the proposed framework, while carefully balancing the additional computational and communication overhead introduced by richer inputs. Another limitation is that we use the Minkowski as the sparse engine, which includes rich features, but is not as efficient as other sparse frameworks. Future work could further improve inference speed by implementing features like cross kernels on higher-speed sparse frameworks such as TorchSparse or Spconv.

\bibliographystyle{IEEEtran}
\bibliography{IEEEexample}










\newpage

 
\vspace{11pt}

\vfill

\end{document}